%% file: fl_evaluation_paper.tex
\documentclass[conference]{IEEEtran}
\IEEEoverridecommandlockouts

\usepackage{cite}
\usepackage{amsmath,amssymb,amsfonts}
\usepackage{algorithmic}
\usepackage{hyperref} 
\usepackage{graphicx}
\usepackage{textcomp}
\usepackage{listings}
\usepackage{svg}
\usepackage{xcolor}
\usepackage{subcaption}

\usepackage[colorinlistoftodos,textsize=scriptsize]{todonotes}

\lstdefinestyle{smallpython}{
    language=Python,
    basicstyle=\footnotesize\ttfamily,
    showstringspaces=false,
    framerule=0.4pt,
    breaklines=true,
    framesep=3pt,
    aboveskip=6pt,
    belowskip=4pt
}

\def\BibTeX{{\rm B\kern-.05em{\sc i\kern-.025em b}\kern-.08em
    T\kern-.1667em\lower.7ex\hbox{E}\kern-.125emX}}
\begin{document}
\title{FLAM: Evaluating Model Performance with Aggregatable Measures in Federated Learning\\
\thanks{This work was co-funded by the Federal Ministry of Research, Technology and Space (BMFTR) under Grant 13FH587KX1 (FederatedForecasts)

© 2026 IEEE. Personal use of this material is permitted. Permission from IEEE must be
obtained for all other uses, in any current or future media, including
reprinting/republishing this material for advertising or promotional purposes, creating new
collective works, for resale or redistribution to servers or lists, or reuse of any copyrighted
component of this work in other works.
}
}

\author{\IEEEauthorblockN{Fabian Stricker}
\IEEEauthorblockA{%
\textit{Institute of Data-Centric}\\ 
\textit{Software Systems (IDSS)},\\
\textit{Hochschule Karlsruhe},\\
Karlsruhe, Germany \\
fabian.stricker@h-ka.de}
\and
\IEEEauthorblockN{Jose A. Peregrina}
\IEEEauthorblockA{%
\textit{Institute of Data-Centric}\\
\textit{Software Systems (IDSS)},\\
\textit{Hochschule Karlsruhe},\\
Karlsruhe, Germany\\
joseantonio.peregrinaperez@h-ka.de}
\and
\IEEEauthorblockN{David Bermbach}
\IEEEauthorblockA{\textit{Scalable Software Systems} \\ \textit{Research Group} \\
\textit{TU Berlin}\\
Berlin, Germany \\
db@3s.tu-berlin.de}
\and
\IEEEauthorblockN{Christian Zirpins}
\IEEEauthorblockA{
\textit{Institute of Data-Centric}\\ 
\textit{Software Systems (IDSS)},\\
\textit{Hochschule Karlsruhe},\\
Karlsruhe, Germany\\
christian.zirpins@h-ka.de}
}

\maketitle

\begin{abstract}
\input{sections/abstract.tex}
\end{abstract}


\section{Introduction}
\input{sections/introduction.tex}

\section{Related Work}
\input{sections/related_work.tex}

\section{Performance Evaluation in Federated Learning}
\input{sections/performance_evaluation_in_fl.tex}

\section{Aggregation Deviations of Metrics} 
\input{sections/inacurate_aggregation_of_metrics.tex}

\section{Achieving Centralized Evaluation with distributed datasets}
\input{sections/proposed_solution.tex}

\section{Conclusion}
\input{sections/conclusion.tex}

\bibliographystyle{IEEEtran}
\bibliography{auto_export_bib}

\end{document}

%% file: sections/abstract.tex
Performance evaluation is essential for assessing the quality of machine learning (ML) models and guiding deployment decisions.
In federated learning (FL), assessing the performance is challenging because data are distributed across participants. 
Consequently, the coordinator must rely on locally computed evaluation metrics and aggregate them to assess the global model.
A key challenge is that common aggregation strategies, such as weighted averaging based on the local samples per participant, do not always produce the same results as centralized evaluation. Existing definitions of performance evaluation are largely tailored to accuracy and do not generalize to other metrics, leading to inconsistencies between participant-based and centralized evaluation.
However, such discrepancies are inconsistent with the FL objective and lead to a wrong calculation of the metric.

To address this issue, we examine the underlying reasons for these discrepancies and propose FLAM, a performance evaluation method based on aggregatable measures that yields the same results as centralized evaluation without the need for a global test dataset.

%% file: sections/introduction.tex
Federated learning (FL) is an emerging paradigm for enabling the training of a machine learning (ML) model in a federated way across multiple participants.
Each participant trains a ML model locally on their own private dataset and sends the resulting model to a coordinator.
The coordinator then aggregates the models into a global model, which aims to contain all the knowledge of all participants in order to create a model that generalizes well~\cite{mcmahanCommunicationEfficientLearningDeep2017a, hardFederatedLearningMobile2019}.
In many applications, such as healthcare and the energy sector, the data are not publicly available because they contain sensitive information about patients or customers, which cannot be collected centrally due to privacy regulations (e.g., GDPR\footnote{https://gdpr.eu/}).
Using FL, the data can stay at each participant's site because the local model acts as a substitute, which improves both communication efficiency and privacy~\cite{mcmahanCommunicationEfficientLearningDeep2017a}.
When institutions collaborate, it is important to accurately assess the model's performance in order to make correct decisions about deployment, retraining, or changing the model architecture.
Especially in FL, where participants need to coordinate, a wrong decision can lead to increased costs for all parties.
Since the coordinator does not have data, it needs to leverage the participant's data to assess the performance of the model.

Today, there is no standardized approach on how to perform the evaluation across different kinds of metrics and how to interpret the evaluation results despite the existence of a benchmarking framework~\cite{caldasLEAFBenchmarkFederated2019}. 
Furthermore, for the participant-based evaluation, it is necessary to define how to aggregate the evaluation results.
Here, different weights can yield different results, which makes a comparison not meaningful.
In studies, weights based on the number of samples per participant~\cite{laiOortEfficientFederated2021} or uniform weights are used~\cite{choFLAMEFederatedLearning2022a, yuTCTConvexifyingFederated2022}.
In some cases, the metrics are calculated on a coordinator-side dataset~\cite{gargClientAvailabilityFederated2025,sunMimiCCombatingClient2024} without the need for an aggregation.

Besides the selection of weights, we observed a significant difference between participant-based evaluation and sending local datasets to the coordinator for centralized evaluation.
This issue particularly affects the aggregation of performance metrics that follow the implementation of frameworks~\cite{caldasLEAFBenchmarkFederated2019,beutelFlowerFriendlyFederated2022} or definitions of evaluations~\cite{chaiFedEvalHolisticEvaluation2022, chaiSurveyFederatedLearning2024} that use the number of samples as weights.
However, this issue should not occur because the evaluation follows the FL objective~\cite{mcmahanCommunicationEfficientLearningDeep2017a} where each sample has the same importance~\cite{chaiSurveyFederatedLearning2024}, which is also the case for centralized evaluation.
Furthermore, with an increasing non-independent and identically distributed (IID) skew, the differences get more severe.
Resolving these differences is important because they lead to a wrong assessment of the performance.
Furthermore, a solution needs to be generic in order to be usable in combination with many performance metrics.

%
In order to address this problem, we analyze the discrepancy by running experiments with a set of commonly used metrics for classification and regression tasks while considering few participants.
In addition, we discuss the evaluation in FL in order to clarify comparison issues and identify important aspects that need to be considered when evaluating the performance of a model.
To resolve the discrepancy, we propose FLAM, a method to evaluate models in FL with aggregatable measures that leverages the data of participants while achieving the same results as a centralized evaluation.
The method eases the comparison with centralized approaches and supports mechanisms that usually require a centralized test dataset to calculate the performance.
In summary, we make the following contributions:
\begin{enumerate}
    \item We discuss performance evaluation in FL and the aspects that need to be considered (Section~\ref{sec:performance_eval_fl}). 
    \item We analyze whether commonly used metrics in regression and classification tasks can be aggregated using weighted averaging to produce the same results as centralized evaluation. Furthermore, we analyze the aspects that influence the differences (Section~\ref{sec:inacurate_aggregation_of_metrics}).
    \item We propose an evaluation method based on aggregatable measures that leverages the participants to achieve the same result as centralized evaluation (Section~\ref{sec:solution}).
    \item We discuss privacy and applicability aspects of the evaluation method for real-world FL systems (Section~\ref{sec:solution}).
\end{enumerate}

%% file: sections/related_work.tex
\label{sec:related_work}
In this section, we discuss the related work on performance evaluation in FL. Throughout this paper, we use the term evaluation to denote performance evaluation.

In FL, there are a few evaluation frameworks that do not solely focus on evaluating the performance of the model but rather on the FL system with a variety of properties such as communication costs, data heterogeneity, and convergence time~\cite{caldasLEAFBenchmarkFederated2019, huOARFBenchmarkSuite2022,nilssonPerformanceEvaluationFederated2018a,chaiFedEvalHolisticEvaluation2022, liuEvaluationFrameworkLargescale2020}.
In this regard, the frameworks only consider different types of the accuracy metric, which is often insufficient because it does not consider the data distribution.
Consequently, additional metrics should be considered to provide more insights into the quality of the model.

Chai et al.~\cite{chaiSurveyFederatedLearning2024} provide a definition for the evaluation of variety of FL system properties including the model performance, which they refer to as utility. 
They mention that the model trained with FL should ideally achieve the same performance as a centralized trained one. Furthermore, they provide a definition for participant-based evaluation with focus on the accuracy metric. 
Although this paper provides insights into the evaluation, they are only addressing the accuracy metric without considering alternatives. Furthermore, a detailed discussion about which aspects have to be considered is missing.

Besides the definition of the evaluation, some studies mention how to select the weights for the aggregation. 
Here, commonly mentioned weights are either distributed uniformly in order to provide more fairness or weighted by the number of samples to prevent a participant with few samples from having a high influence on the final score~\cite{caldasLEAFBenchmarkFederated2019, chaiSurveyFederatedLearning2024}.

Flach~\cite{flach2019performance} discusses good and bad practices for evaluation in ML to provide an important foundation for a measurement theory. 
In detail, he shows how to aggregate performance metrics for binary classification based on the confusion matrix considering cross-validation.
Furthermore, he mentions that the F1-score should not be aggregated using arithmetic averages because it can lead to incoherent results.
This is interesting because we can identify more severe issues in FL.
In contrast, FL is subject to heterogeneous data, where participants can have a vastly different label distribution and quantity. 
Besides data, privacy is an important aspect that needs to be considered when designing an approach.

Our paper differs from the related work because we focus on the evaluation of FL trained models to achieve the same result as centralized evaluation. 
Furthermore, we focus on aspects unique to FL, while discussing the goal of the evaluation.
In addition, we do not solely focus on classification tasks but aim for a generalizable approach.
To the best of our knowledge, there is no paper that addresses the identical issue for heterogeneous FL settings.

%% file: sections/performance_evaluation_in_fl.tex
\label{sec:performance_eval_fl}
In this section, we provide an overview of evaluation in FL and clarify which aspects need to be considered when evaluating the model performance. 
%

In prominent frameworks such as Flower and NVIDIA Flare~\cite{beutelFlowerFriendlyFederated2022,rothNVIDIAFLAREFederated2022} multiple evaluation methods are mentioned, from coordinator-side evaluation with a test set on the coordinator and participant-based evaluation to cross-site evaluation, where all global and local models are evaluated on all participants.
Here, we only consider the centralized evaluation and the participant-based evaluation because they do not reveal the participant model to other participants. %
Furthermore, we consider supervised ML tasks and the evaluation of the global model.
We first define the coordinator-side evaluation based on the Flower implementation~\cite{beutelFlowerFriendlyFederated2022} which can be denoted as Eq.~\ref{eq:server_based_evaluation} that partially aligns with the definition of centralized effectiveness of Chai et al.\cite{chaiSurveyFederatedLearning2024} excluding the centralized training:

\begin{equation}
\label{eq:server_based_evaluation}
Utility(Predict(w_g, D_x), D_y)
\end{equation}
In the equation, $Utility()$ denotes a function that calculates a performance metric and the function $Predict()$ depicts the prediction made by the global model $w_g$ based on all samples, where $D_x$ are the features and $D_y$ the assigned labels per sample. 

Next, we look at the definition of participant-based evaluation.
For the definition, we follow the Flower~\cite{beutelFlowerFriendlyFederated2022} implementation and the definition of Chai et al.\cite{chaiSurveyFederatedLearning2024}. 
Although the implementation is generic for any metric, the definition describes only the accuracy metric.
We combine both aspects and get the following definition: 
\begin{equation}
\label{eq:client_based_eval_preliminary}
\sum_{i=1}^{P} \lambda_i * Utility(Predict(w_g, D_{i_x}), D_{i_y})
\end{equation}
This definition describes that the evaluation result is calculated following the weighted average of a utility function for a set of  participants, where $P$ refers to the number of participants. 
Here, $\lambda_i$ refers to the weight assigned to the participant $i$, which can be, for instance, computed based on the proportion of data samples for a participant $|D_i|$ relative to the samples of the global dataset $|D_{all}|$, which is the concatenation of all participant datasets. 
If we select these weights, then the definition aligns with the FL objective~\cite{mcmahanCommunicationEfficientLearningDeep2017a}. Furthermore, $D_{i_x}$ denotes the features for the samples of the participant $i$.
The resulting definition faces two issues.
First, it is not defined how the weights should be selected, as some studies use the number of samples as weights~\cite{laiOortEfficientFederated2021} and others use uniform weights~\cite{yuTCTConvexifyingFederated2022, choFLAMEFederatedLearning2022a}.
Second, different weights can lead to divergent interpretations, which, in turn, can result in incoherent metrics if it is not described how they were calculated.
%
To this end, we want to discuss which aspects are important for evaluation as well as how to handle different computations of metrics.

\noindent\textbf{Which aspects are important for evaluation?}
In order to discuss this question, we start with evaluation in centralized ML.
The evaluation metrics are selected based on the properties of the available dataset (e.g., class imbalance) as well as the learning task (e.g., regression, classification) and the loss function, which can be modified to match the learning task (e.g., the cost of different predictions or regularization).
In comparison, FL introduces various new properties and changes in the learning goal that lead to new aspects that have to be considered during evaluation.
First, the selection of the weights for the aggregation of the participant metrics is important because the weights modify the resulting metric. 
Chai et al.~\cite{chaiSurveyFederatedLearning2024} recommend using the number of samples per participant as weight because they follow the loss of the optimization goal defined by McMahan et al.~\cite{mcmahanCommunicationEfficientLearningDeep2017a} which defines that each sample gets assigned the same importance.
However, in research there are aggregation methods that modify the FL objective~\cite{liDittoFairRobust2021a,liFairResourceAllocation2020a}.
Therefore, the weight selection depends on the objective.
If they do not align with the objective, it could lead to a model that achieved the optimal minimum, but the metric does not represent it. Consequently, one could make wrong assumptions about the quality of the model.

Second, when selecting metrics for evaluating the performance, it is unknown whether the dataset has class imbalance because the data are private and distributed across the participants.
In this case, plain accuracy can be misleading because the model can optimize on a class that occurs often.
Hence, it would be better to use multiple metrics or metrics that can handle both cases to prevent a wrong assessment.

Third, in FL the data can change at any time due to new data collected by the participants or changes in the federation.
For example, participants can join, leave, or encounter failures, which leads to a permanent or temporary change in data distribution that can lead to a different quality assessment~\cite{strickerAnalyzingImpactParticipant2025b}.
Hence, considering the generalization over the quality for a fixed dataset can be more important.

Fourth, the model is not created for a single entity but rather for a federation of participants.
For example, if a low number of market competitors collaborate, the evaluation needs to consider fairness so that all participants benefit from participation; otherwise, they may not participate at all~\cite{kuoResearchCollaborativeLearning2025}.

Last, in FL we do not have one resulting model but rather multiple models, such as the local models, the global model, or multiple personalized versions of the global model~\cite{liDittoFairRobust2021a}.
Here, it is important to define what should be evaluated and which goals the evaluation should follow. 
It would be reasonable to evaluate the final model that is used as a service and the global model, which can be considered an intermediate result.
Consequently, the calculation and selection of the metrics need to align with the goal of the FL task and need to consider unique FL aspects.

\noindent\textbf{How can we differentiate and compare interpretations?}
If we consider the important aspects for evaluation, then the calculation of metrics is complex, and denoting the name of the metric is insufficient, because the aggregation changes the measurement, which can lead to wrong conclusions and comparisons.
Consequently, it is necessary that studies mention how the metrics are aggregated and what they measure.
Although a comparison between different measurements is not meaningful, they could be used as extensions to provide supporting insights about the quality of the model.
For example, assigning uniform weights to each participant can yield more information about the fairness of the process~\cite{liFairResourceAllocation2020a}. 
If we want to compare the participant-based evaluation with the centralized ML evaluation, then it is necessary to understand how the weights should be selected. 
In centralized evaluation, every sample counts equally, which is also the case for the FL optimization objective~\cite{mcmahanCommunicationEfficientLearningDeep2017a}.
Consequently, a comparison of these two metrics  should lead to a meaningful evaluation because they refer to the same measurement.
Contrary to the expectations, some metrics lead to different results between centralized evaluation and aggregating them based on the weighted average.
This can lead to wrong conclusions when comparing centralized training against federated training and evaluation.

%% file: sections/inacurate_aggregation_of_metrics.tex
\label{sec:inacurate_aggregation_of_metrics}
In this section, we show that computing the metric using the number of samples per participant as weights in a weighted averaging aggregation can yield different results compared to coordinator-side evaluation.
Furthermore, we identify for which metrics the deviations occur and what factors lead to large deviations.
We first define the experiment setup and then analyze the experiments.
\subsection{Experiment Setup}
To demonstrate the severity of the effect and underscore the relevance in federated settings, we conducted experiments across multiple datasets and learning tasks.
In Table~\ref{tab:exp_configuration}, we summarize the configuration of the experiments.
\begin{table}[!t]
\caption{Experiment Configurations}
\centering
\begin{tabular}{|c|c|}
\hline
\textbf{Property} & \textbf{Value} \\
\hline
Datasets (Classification) & Covertype, CIFAR-10, CIFAR-100 \\
Metrics (Classification) & Accuracy, Precision, Recall, F1-score, MCC\\
\hline
Datasets (Regression) & PVOD, GermanSolarFarm \\
Metrics (Regression) & R2-score \\
\hline
Aggregation Algorithm & FedAvg~\cite{mcmahanCommunicationEfficientLearningDeep2017a} \\
Number of participants & 4 \\
Local Training Rounds & 5 \\
Communication Rounds & 20 \\
Optimizer & Adam \\
\hline
\end{tabular}
\label{tab:exp_configuration}
\end{table}
For the datasets, we selected the CIFAR-10~\cite{krizhevskyLearningMultipleLayers2009a} and CIFAR-100~\cite{krizhevskyLearningMultipleLayers2009a} for image classification because they are balanced and feature different complexities with a high number of classes.
In the case of the tabular datasets, we select the Covertype~\cite{blackardCovertype1998} dataset because it contains many features and class imbalance.
Last, the time-series datasets PVOD~\cite{yaoPVODV10Photovoltaic2021} and GermanSolarFarm~\cite{genslerGermanSolarFarmDataSet2016} are selected because they feature a different learning task and contain heterogeneity through the different photovoltaic stations.

For the partitioning of the classification datasets, we use the Dirichlet distribution to generate different data distributions, as it is commonly used in FL\cite{wangTacklingObjectiveInconsistency2020a,linEnsembleDistillationRobust2020}. 
In order to artificially create quantity skews (QS), label skews (LS) or a combination of quantity and label skews (LQS), we use the alpha values in Table~\ref{tab:alpha_values}.
In addition to the Dirichlet-based skews, we create a manual skew (MS). Here, participants share few classes but also are the sole owner of a specific class, which simulates a scenario where organizations have partially overlapping distributions but also maintain unique edge cases.
\begin{table}[!t]
\caption{Alpha Parameters used for Data Partitioning}
\centering
\begin{tabular}{|c|c|}
\hline
\textbf{Property} & \textbf{Value}  \\
\hline
QS Alpha Values & 2.0 \\
LS Alpha Values & 0.6, 7.0 \\
LQS Alpha Values (QS, LS) & (10.0, 1.0), (3.0, 0.7)  \\ 
\hline
\end{tabular}
\label{tab:alpha_values}
\end{table}
In the case of the time-series dataset, we select two forecasting datasets that consist of multiple stations.
For the PVOD dataset, we choose 4 stations that differ in the number of observations.
In the case of the GermanSolarFarm, we group the stations together and assign each participant one group to simulate energy providers with different stations located across Germany.

Next, we explain the model types used for each dataset.
For both CIFAR-10 and CIFAR-100, we use convolutional neural networks (CNNs) in combination with dense layers for the output to achieve reasonable performance with acceptable computational overhead.
In case of the Covertype dataset, we use a TabTransformer~\cite{huangTabTransformerTabularData2020} because they achieve good performance in few training rounds.
For the time series datasets, we use a one-dimensional CNN in combination with gated recurrent units (GRU) for the PVOD dataset, as these provide high quality with low computational resources. For the GermanSolarFarm dataset, we use a long short-term memory (LSTM) model.
For all implementations, we use Tensorflow\footnote{https://www.tensorflow.org/} and calculate the metrics using Scikit-learn\footnote{https://scikit-learn.org/stable/}.

Based on these configurations, we run experiments for each dataset that results from the partitioning. 
Here, the goal is to identify if there are classification or regression metrics that show deviations between the coordinator-based evaluation on the concatenated participant datasets and participant-based evaluation.
As metrics, we selected commonly used metrics in multiclass classification, such as accuracy and the macro and weighted versions of precision, recall, and F1-score. 
Furthermore, we added the Matthews Correlation Coefficient (MCC)\cite{gorodkinComparingTwoKcategory2004} as a balanced measure even if the datasets have  class imbalance~\cite{jurmanComparisonMCCCEN2012}.
For the regression tasks, we selected the R2-score, which measures the proportion of the variance of a target that can be explained by the model.

\subsection{Findings}
\begin{figure}[!t]
\centering
\includegraphics[width=195px]{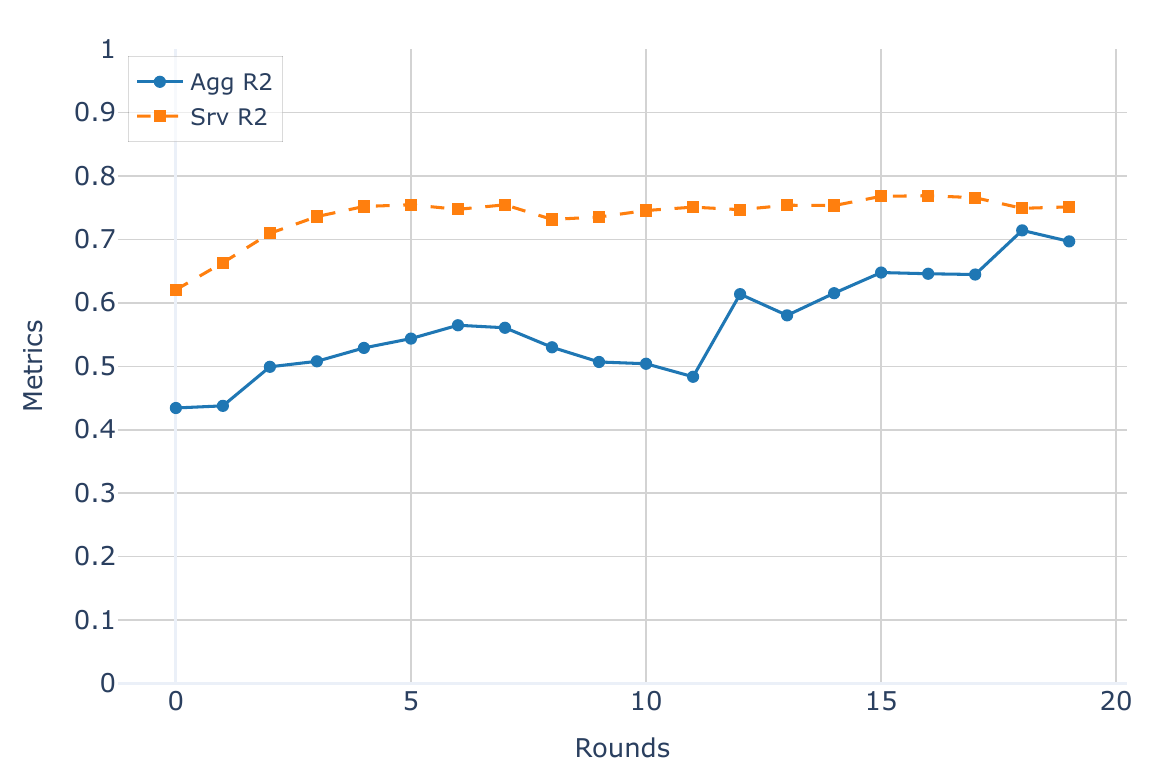}
\caption{PVOD: Comparison of aggregated metrics against coordinator-side calculated metrics.}
\label{fig:prem_exp_pvod_comp_aggregated_metrics_vs_server_side}
\end{figure}

\begin{figure}[!t]
\centering
\begin{subfigure}[t]{195px}
\centering
\includegraphics[width=195px]{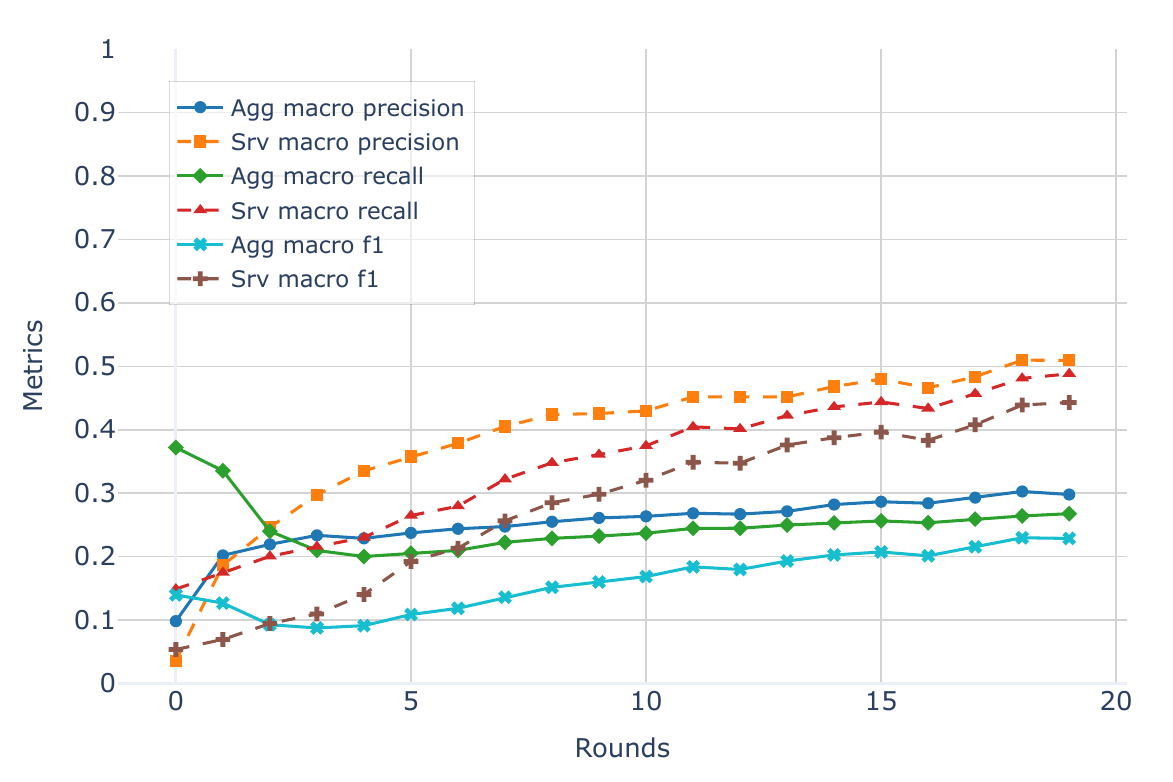}
\caption{Macro Metrics}
\label{fig:prem_exp_cifar_10_ms_mg1}
\end{subfigure}
\begin{subfigure}[t]{195px}
\centering
\includegraphics[width=195px]{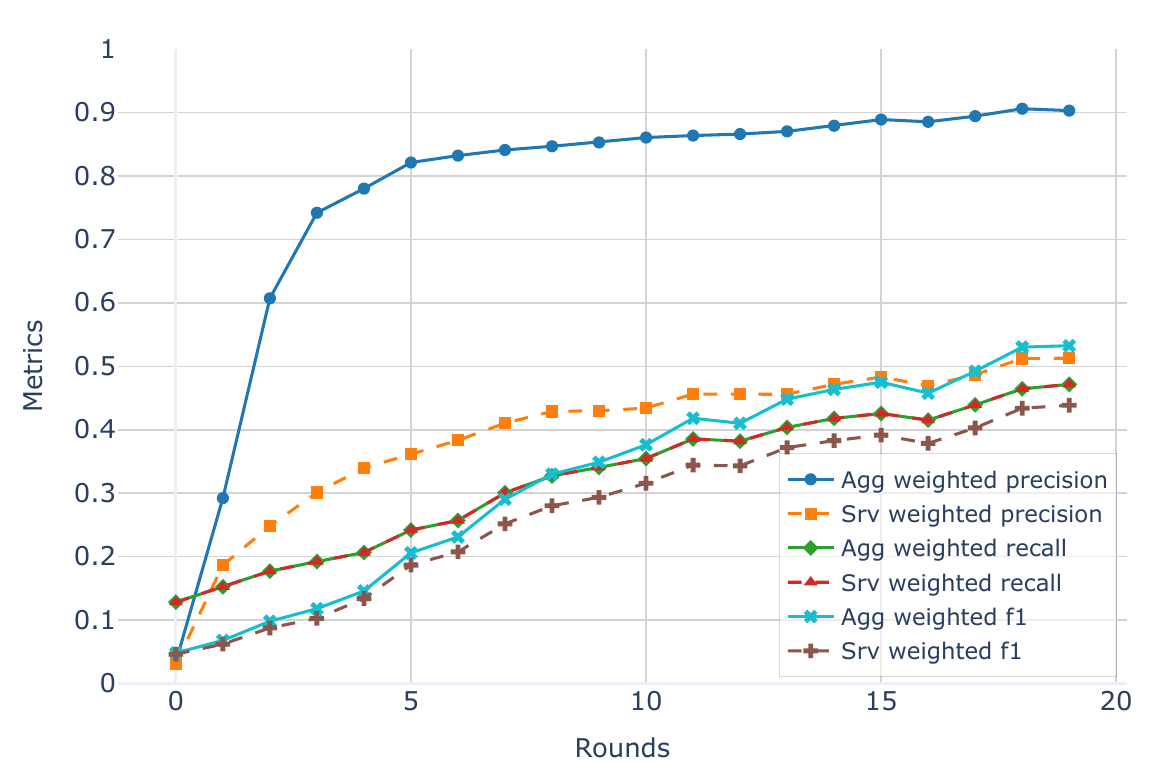}
\caption{Weighted Metrics}
\label{fig:prem_exp_cifar_10_ms_mg2}
\end{subfigure}
\begin{subfigure}[t]{195px}
\centering
\includegraphics[width=195px]{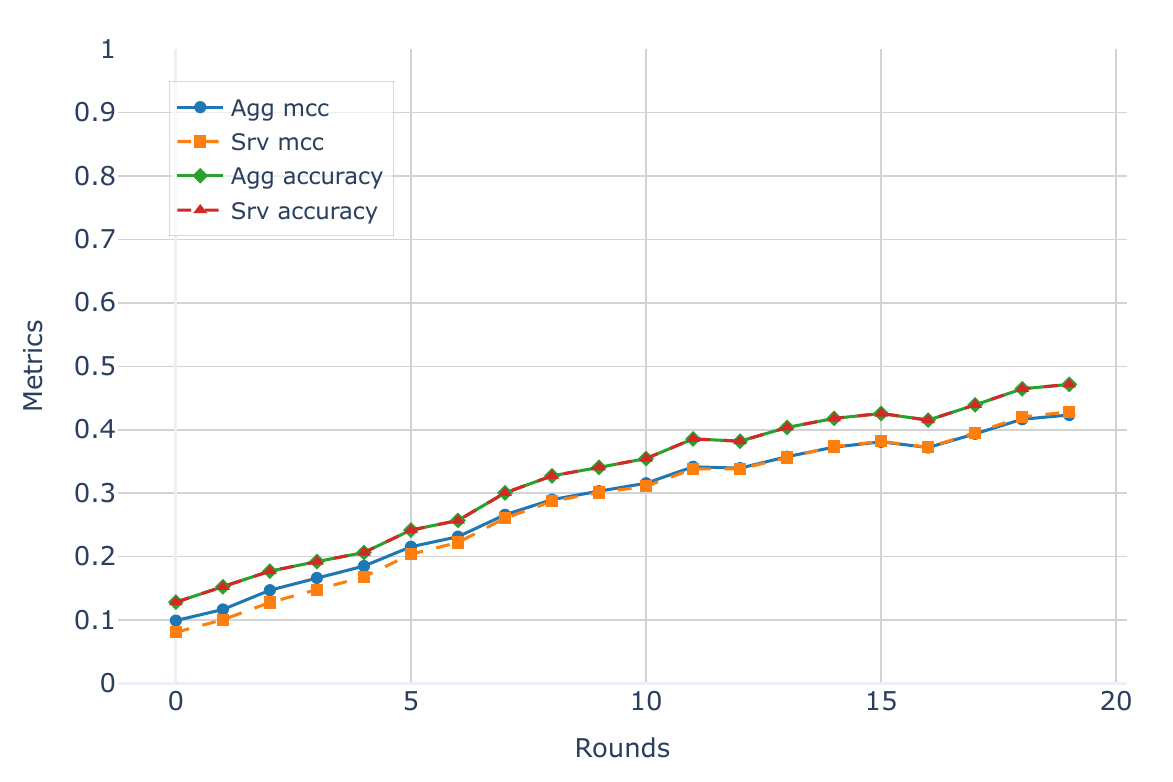}
\caption{MCC and Accuracy}
\label{fig:prem_exp_cifar_10_ms_mg3}
\end{subfigure}
\caption{CIFAR-10 MS: Comparison of aggregated metrics against coordinator-side calculated metrics.}
\label{fig:prem_exp_cifar10_ms_combined}
\end{figure}

Following the experiment setup, we conduct multiple experiments and summarize the following observations:

\noindent\textbf{Finding 1: The aggregation of metrics such as the R2-score, Macro F1-score, Macro Precision, Macro Recall, Weighted F1-score, MCC and Weighted Precision showed differences compared to a centralized calculation --}
In Fig.~\ref{fig:prem_exp_pvod_comp_aggregated_metrics_vs_server_side}, the R2-score shows a difference of up to 0.26 between aggregating the participant metrics and calculating the results on a coordinator-side dataset.
For the CIFAR-10 dataset in Fig.~\ref{fig:prem_exp_cifar10_ms_combined}, we can also observe differences across the metrics.
Here, the weighted precision shows the highest differences, with up to 0.39. 
In comparison, the weighted recall and accuracy show no differences and have the same results as centralized evaluation.

\begin{figure}[!t]
\centering
\begin{subfigure}[t]{195px}
    \centering
    \includegraphics[width=195px]{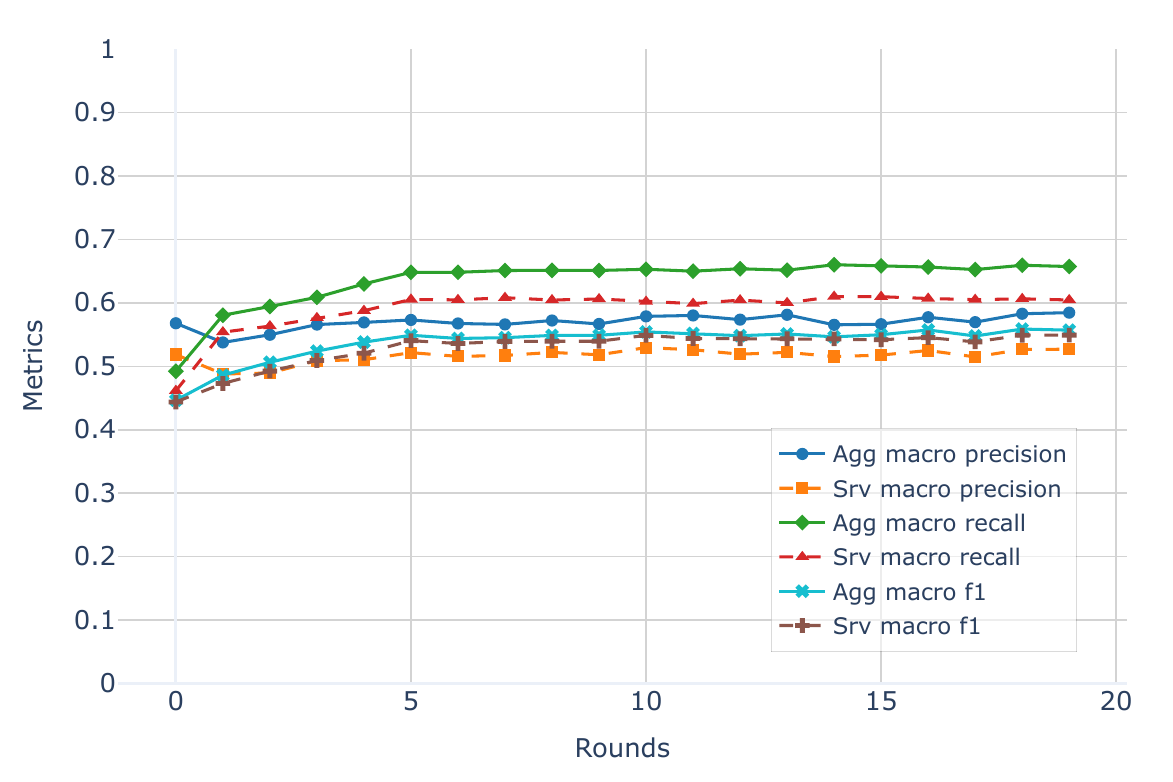}
    \caption{Macro Metrics}
    \label{fig:prem_exp_covertype_ms_mg1}
\end{subfigure}
\begin{subfigure}[t]{195px}
  \centering
  \includegraphics[width=195px]{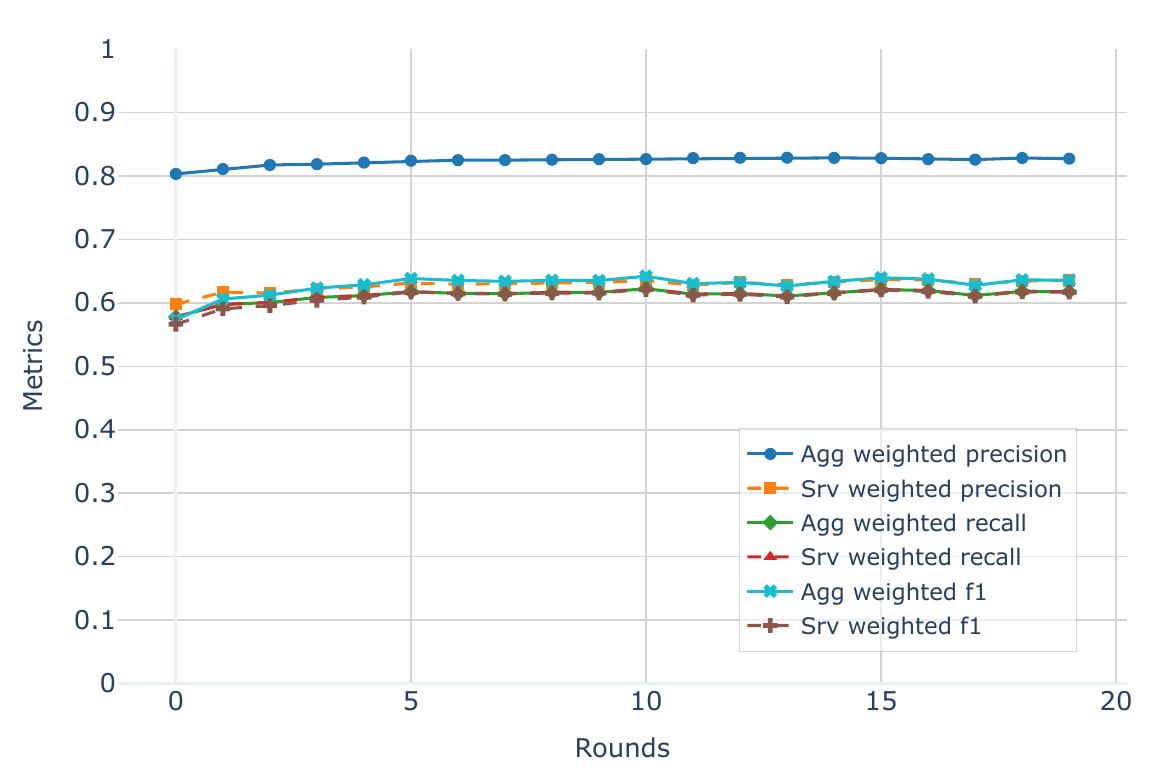}
  \caption{Weighted Metrics}
  \label{fig:prem_exp_covertype_ms_mg2}
\end{subfigure}
\begin{subfigure}[t]{195px}
  \centering
  \includegraphics[width=195px]{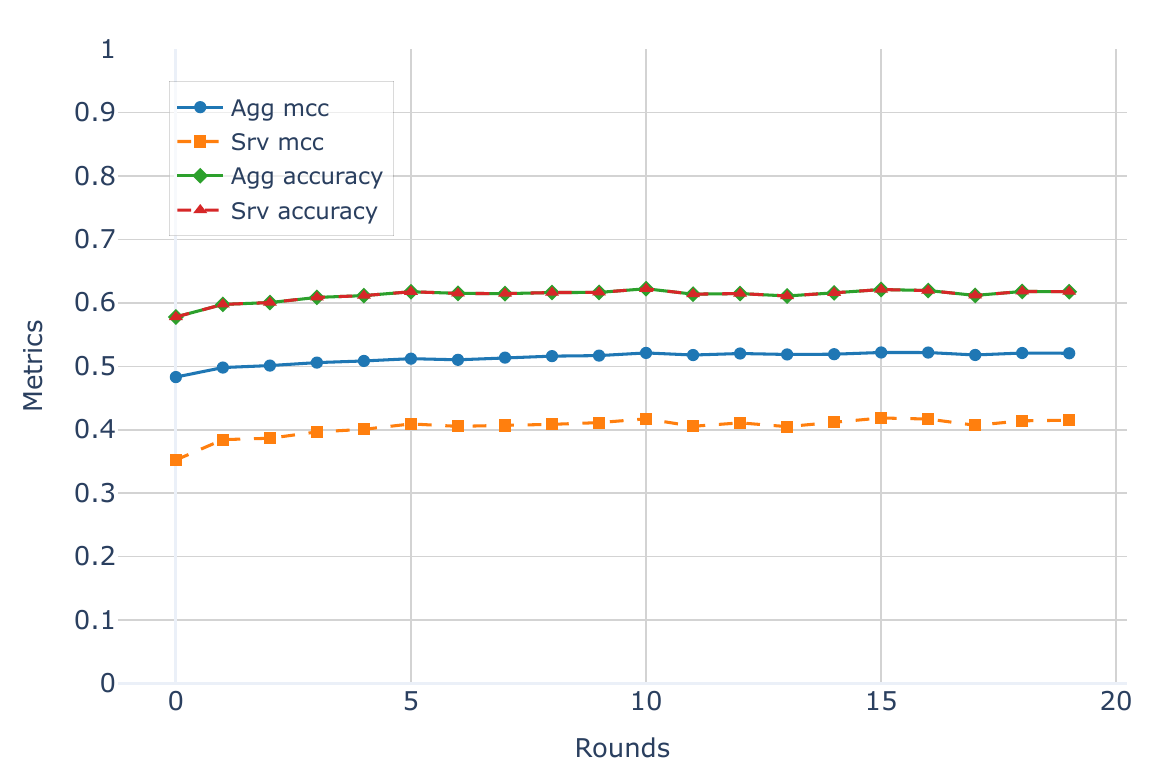}
  \caption{MCC and Accuracy}
  \label{fig:prem_exp_covertype_ms_mg3}
\end{subfigure}
\caption{Covertype LS with $\alpha=0.6$: Comparison of aggregated metrics against coordinator-side calculated metrics.}
\label{fig:prem_exp_covertype_ms_combined}
\end{figure}

In addition, we can observe differences for the Covertype dataset across the different metrics in Fig.~\ref{fig:prem_exp_covertype_ms_mg1} and ~\ref{fig:prem_exp_covertype_ms_mg2}. 
Although in Fig.~\ref{fig:prem_exp_covertype_ms_mg3} the MCC indicates larger differences than for the CIFAR10 dataset. 
Here, the accuracy and weighted recall both show no differences and are similar to coordinator-side evaluation. 
The highest absolute difference was for the weighted precision, with up to 0.19.
Based on the experiments, we showed that there can be differences between aggregating the metrics and calculating them based on a coordinator-side dataset.
Upon analyzing the different skews of the experiments, we could observe that the differences are higher the larger the deviation between the participants' metrics are.

\noindent\textbf{Finding 2: The aggregation of metrics such as the accuracy and the weighted recall showed no deviations --}
In the previous finding, we could observe differences for most metrics except accuracy and weighted recall.
Rather than showing a difference between the evaluations, we want to show that there is no difference.
We assume that the coordinator-side dataset is a concatenation of the participant datasets and use Eq.~\ref{eq:client_based_eval_preliminary}, where the weights for a participant are the number of samples.
We start by analyzing the accuracy metric for multiclass classification and set the centralized calculation equal to the aggregation.
In the Eq.~\ref{eq:accuracy_eq}, the variable $C$ denotes the number of classes and $TP_{i_j}$ refers to the true positives for the class $j$ across the dataset for participant $i$.
\begin{align}
\label{eq:accuracy_eq}
\frac{\sum_{j=1}^{C} TP_{{all}_j}}{|D_{all}|} &= \sum_{i=1}^{P} \textstyle \frac{|D_i|}{|D_{all}|} * \frac{\sum_{j=1}^{C} TP_{{i}_j}}{|D_i|} \notag \\
  &= \frac{1}{|D_{all}|} \sum_{i=1}^{P} \sum_{j=1}^{C} TP_{{i}_j} \notag \\
  &= \frac{\sum_{j=1}^{C} TP_{{all}_j}}{|D_{all}|}
\end{align}
Furthermore, through simplification, we show that the centralized calculation of accuracy is equal to the aggregation.
Following the same approach, we can show the same behavior for the weighted recall in Eq.~\ref{eq:w_recall_eq}. 
Here, we use the formula from Scikit-learn\cite{34MetricsScoring} that uses the number of samples for a class as weights. 
The left-hand side represents the centralized evaluation and is already simplified for clarity. 
We show equivalency by simplifying the right-hand side which describes the weighted recall with aggregation.
\begin{align}
\label{eq:w_recall_eq}
  \frac{\sum_{j=1}^{C} TP_{{all}_j}}{|D_{all}|} &=\sum_{i=1}^{P} \textstyle \frac{|D_{i}|}{|D_{all}|} \frac{1}{|D_{i}|}  \sum_{j=1}^{C} \frac{TP_{i_j} + FN_{i_j}}{1}\frac{TP_{i_j}}{TP_{i_j} + FN_{i_j}} \notag \\
  &= \frac{1}{|D_{all}|} \sum_{i=1}^{P} \sum_{j=1}^{C} TP_{{i}_j} \notag \\
  &= \frac{\sum_{j=1}^{C} TP_{{all}_j}}{|D_{all}|} 
\end{align}
Upon simplification, we observe that in the context of multiclass classification, the weighted recall with samples per class as weights leads to the same measurement as the accuracy.
The selection of the weights is crucial because they modify the equation to measure the number of all TPs against all samples, which is equivalent to accuracy.

Following both equations, we demonstrated that it is possible to achieve the same results as coordinator-side evaluation in a distributed way, leveraging the participants.
As a consequence, the predictions are not the cause of the difference; otherwise, all metrics have to be affected.
However, we showed that only a subset of metrics are affected.
Leading to the conclusion that there is no difference between making predictions with a fixed global model in a centralized location versus making the predictions with the same model in a distributed manner.
Hence, there is no dependency between input samples.
This conclusion is only true if we consider our assumptions; hence, the global model on each participant is fixed and deterministic. Therefore, the prediction for the same sample will yield the same result because the prediction does not modify the model weights.
Consequently, the distributed evaluation process can be interpreted as parallelization of the evaluation.
To resolve the problem of the different values, a solution needs to modify the calculation of the metrics, which we propose in the next section.

%% file: sections/proposed_solution.tex
\label{sec:solution}
In this section, we propose FLAM, a method that calculates performance metrics based on aggregatable measures (AM).
We first define the requirements that the AMs must satisfy.
Next, we demonstrate, given a set of AMs, the calculation is valid.
We then derive the AMs for the R2-score and the F1-scores and show their correctness.
Furthermore, we discuss the applicability and privacy of the proposed method.

\subsection{FLAM: Metric Calculation based on Aggregatable Measures}
Rather than calculating the metrics on the participants side and aggregating them on the coordinator's, we move the calculation of the metrics towards the coordinator.
Thus, we can avoid the aggregation of the performance metrics.
Instead, we identify AMs of the metric calculation and send them to the server to calculate the performance metric.
In this context, AMs are decomposed components of a specific metric, which can be recombined to achieve the same calculation process. 
Hence, they do not modify the calculation of the metric.
We define an AM as a measure that can be calculated on different local datasets and achieves the same results as if they were calculated centrally.
Consequently, an AM needs to satisfy Eq.~\ref{eq:linear_components_requirement}.
\begin{equation}
\label{eq:linear_components_requirement}
AM_{all} = \sum_{i=1}^{P} AM_{i}
\end{equation}
Here, we assume that the global dataset is a concatenation of the test datasets used for evaluation.
Some metrics, such as the R2-score require statistical information such as the mean over the labels.
To fulfill the requirement, the statistical information needs to be calculated across the participant datasets and then aggregated into a global statistical metric.
Otherwise, this would lead to dependencies on the local datasets.
If the metric calculation can be decomposed into a set of AMs, we can evaluate the performance of a model with the following steps:
\begin{enumerate}
    \item The participants calculate the measures on their local dataset.
    \item The local AMs are sent to the coordinator.
    \item The coordinator aggregates the measures.
    \item The coordinator uses the aggregated measures to calculate the metric.
\end{enumerate}
Our method modifies the participant-based evaluation (see Eq.~\ref{eq:client_based_eval_preliminary}) by moving the aggregation of the participant results into the calculation of the metric and removing the weights, because the AMs are calculated based on the samples for a participant.
This results in Eq.~\ref{eq:sub_component_based_calculation}.
In the equation, the function $calc\_AMs_i()$ denotes the calculation of the AMs for a participant $i$ based on the predicted value and the ground truth over the participant dataset $D_i$.
For readability, we shorten the function to the calculated AMs for a participant, assuming that a valid set of AMs is found.
\begin{equation}
\label{eq:sub_component_based_calculation}
\begin{aligned}
Utility(\sum_{i=1}^{P} calc\_AMs_{i}(Predict(w_g, D_{i_x}), D_{i_y})) \\
= Utility(\sum_{i=1}^{P} AM_i)
\end{aligned}
\end{equation}

Now that we have defined the method, we can prove that our method achieves the same performance as centralized evaluation by showing that it is equal to centralized evaluation (see Eq.~\ref{eq:server_based_evaluation}).
In Eq.~\ref{eq:proof_of_method}, we simplify both methods to show that they refer to the same calculation.
To achieve the equivalency, we perform two steps.
First, we rewrite the centralized evaluation to calculate the AMs centrally, which is possible because the AMs are decomposed measures that do not modify the calculation. Hence, AMs are intermediate results of the original metric calculation.
Second, the AMs need to satisfy Eq.~\ref{eq:linear_components_requirement}.
Therefore, we can rewrite the sum of the values as the result over the full dataset.
\begin{equation}
\label{eq:proof_of_method}
\begin{aligned}
Utility(Predict(w_g, D_{{all}_x}), D_y) = Utility(\sum_{i=1}^{P} AM_i)\\
Utility(AM_{all}) = Utility(\sum_{i=1}^{P} AM_i)\\
= Utility(AM_{all})\\
\end{aligned}
\end{equation}

Thus, we can use the method if we decompose a specific metric into measures that are aggregatable.
We showed that some metrics lead to wrong results; therefore, we analyze these metrics and decompose them in AMs.
Since each metric is calculated differently, it is not feasible to show the AMs for all identified metrics.
Therefore, we exemplarily show the method for the R2-score and the F1-scores because they are widely used.
We make the implementations and selection of AMs for all identified metrics including the MCC, weighted precision, macro recall, and macro precision publicly available as a GitHub repository\footnote{https://github.com/HKA-IDSS/Supplement-FLAM-Evaluation}.

Although deriving a general method that identifies a set of AMs is challenging, we propose a structured approach to identify AMs. 
First, we analyze the formal definition of the metric and identify dependencies on dataset-specific statistics. 
For example, quantities such as the mean of a local data partition or other participant-specific scaling factors that can influence the computation.
Second, once a set of AMs has been identified, we validate the correctness by analyzing whether the requirement (see Eq.~\ref{eq:linear_components_requirement}) is satisfied.

We start with finding AMs for the R2-score.
Here, we first analyze the R2-score~\cite{kutnerAppliedLinearRegression2004}:
\begin{equation}
\label{eq:r2_equation}
\begin{aligned}
R^2 = 1 - \frac{\sum_{k=1}^{D_{all}} (y\_true_k - y\_pred_k)^2} {\sum_{k=1}^{D_{all}} (y\_true_k - mean(y\_true_{D_{all}}))^2} 
\end{aligned}
\end{equation}
Based on the equation, we can derive the AMs in Eq.~\ref{eq:r2_comps} and the calculation of the metric on the coordinator, including the placeholders for the AMs in Eq.~\ref{eq:r2_score_components}.
In both equations, $y\_true$ refers to the ground truth and $y\_pred$ to the predicted value for a sample.
\begin{equation}
\label{eq:r2_comps}
\begin{aligned}
rs\_A_{i} = \sum_{k=1}^{D_i} (y\_true_k - y\_pred_k)^2, \\
rs\_B_{i} = \sum_{k=1}^{D_i} (y\_true_k - mean(y\_true_{D_{all}}))^2
\end{aligned}
\end{equation}
\begin{equation}
\label{eq:r2_score_components}
R^2 = 1 - \frac{\sum_{i=1}^{P} rs\_A_{i}}{\sum_{i=1}^{P}rs\_B_{i}} 
\end{equation}
Furthermore, in order to make the calculation identical across all participants, it is necessary to calculate the mean of the concatenated global dataset $D_{all}$. 
Otherwise, this would lead to different results per participant.
To validate the AMs, we start with $rs\_A_{i}$ in Eq.~\ref{eq:r2_comp_A} and verify that the requirement is satisfied.
We can rewrite the double summations as a single one over all samples in the global dataset. 
This is possible because we assume the same deterministic global model across the participants, and the coordinator dataset is the concatenation of the participant datasets. 
Therefore, we have the same prediction per participant for a given sample.
Furthermore, there are no scaling factors connected to the local data part; hence, the requirement is satisfied.
\begin{equation}
\label{eq:r2_comp_A}
\begin{aligned}
\sum_{k=1}^{D_{all}} (y\_true_k - y\_pred_k)^2= \sum_{i=1}^{P} \sum_{k=1}^{D_i} (y\_true_k - y\_pred_k)^2 \\
= \sum_{k=1}^{D_{all}} (y\_true_k - y\_pred_k)^2
\end{aligned}
\end{equation}
We can show the same for the second AM because they have a similar structure. 
Here, the difference is the mean, which needs to be calculated on the full datasets to prevent local dependencies.

Next, we analyze the equation for macro F1-score~\cite{hinojosaleePerformanceMetricsMultilabel2024}:
\begin{equation}
\label{eq:f1_score_macro}
\begin{aligned}
F1\_Macro &= \frac{1}{C} \sum_{j=1}^{C} \frac{2*TP_j}{2*TP_j + FP_j + FN_j}
\end{aligned}
\end{equation}
We can identify the two AMs in \ref{eq:f1_score_comp_macro}, which are the TP for each class as well as the sum of the FP and FN per class.
\begin{equation}
\label{eq:f1_score_comp_macro}
\begin{aligned}
f1\_A_{j_i} &= FP_{i_j} + FN_{i_j}, \\
f1\_B_{j_i} &= TP_{i_j}
\end{aligned}
\end{equation}
Following the AMs, we can rewrite the macro F1 and sum up the AMs across the participants to get Eq.~\ref{eq:f1_macro}.
\begin{equation}
\label{eq:f1_macro}
f1\_m = \frac{1}{C} \sum_{j=1}^{C} \frac{2*\sum_{i=1}^{P} f1\_B_{j_i}}{2*\sum_{i=1}^{P}f1\_B_{j_i} + \sum_{i=1}^{P}f1\_A_{j_i}}
\end{equation}
Next, we repeat the same steps for the weighted F1~\cite{hinojosaleePerformanceMetricsMultilabel2024}.
Here, the equation is modified by using weights based on the number of samples per class instead of uniform weights.
In order to get the weights per class, it is necessary to define another AM that describes the number of samples for a specific class denoted in Eq.~\ref{eq:f1_score_comp_weigthed}.
\begin{equation}
\label{eq:f1_score_comp_weigthed}
w_{j_i} = TP_{i_j} + FN_{i_j}
\end{equation}
Based on this measure, we can modify the weighted F1 to achieve Eq.~\ref{eq:f1_weighted}.
\begin{equation}
\label{eq:f1_weighted}
f1\_w = \frac{1}{D_{all}} \sum_{j=1}^{C} (\sum_{i=1}^{P} w_{j_i}) \frac{2*\sum_{i=1}^{P} f1\_B_{j_i}}{2*\sum_{i=1}^{P}f1\_B_{j_i} + \sum_{i=1}^{P}f1\_A_{j_i}}
\end{equation}
To validate that the AMs for the different F1-scores are correct, we analyze if the requirement is fulfilled.
Since all AMs consist of absolute counts and sums, there is no statistical information or other scaling factor that depends on the local dataset. Consequently, we can perform the same simplification as for the first AM of the R2-score and follow the same reasoning.

\subsection{Results}
\begin{figure}[!t]
\centering
\includegraphics[width=195px]{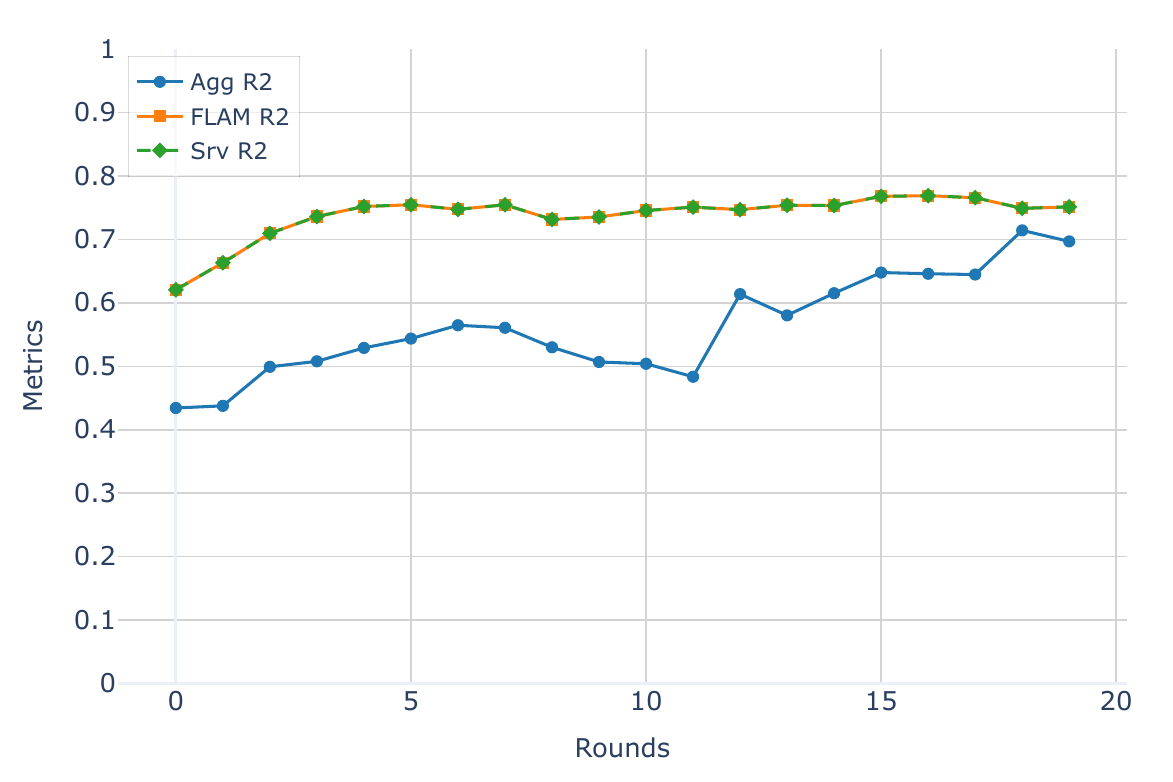}
\caption{PVOD: Comparison of FLAM, coordinator-side calculated metrics, and the aggregated metrics.}
\label{fig:solution_exp_pvod_ms_mg1}
\end{figure}

\begin{figure}[!t]
\centering
\begin{subfigure}{195px}
\includegraphics[width=195px]{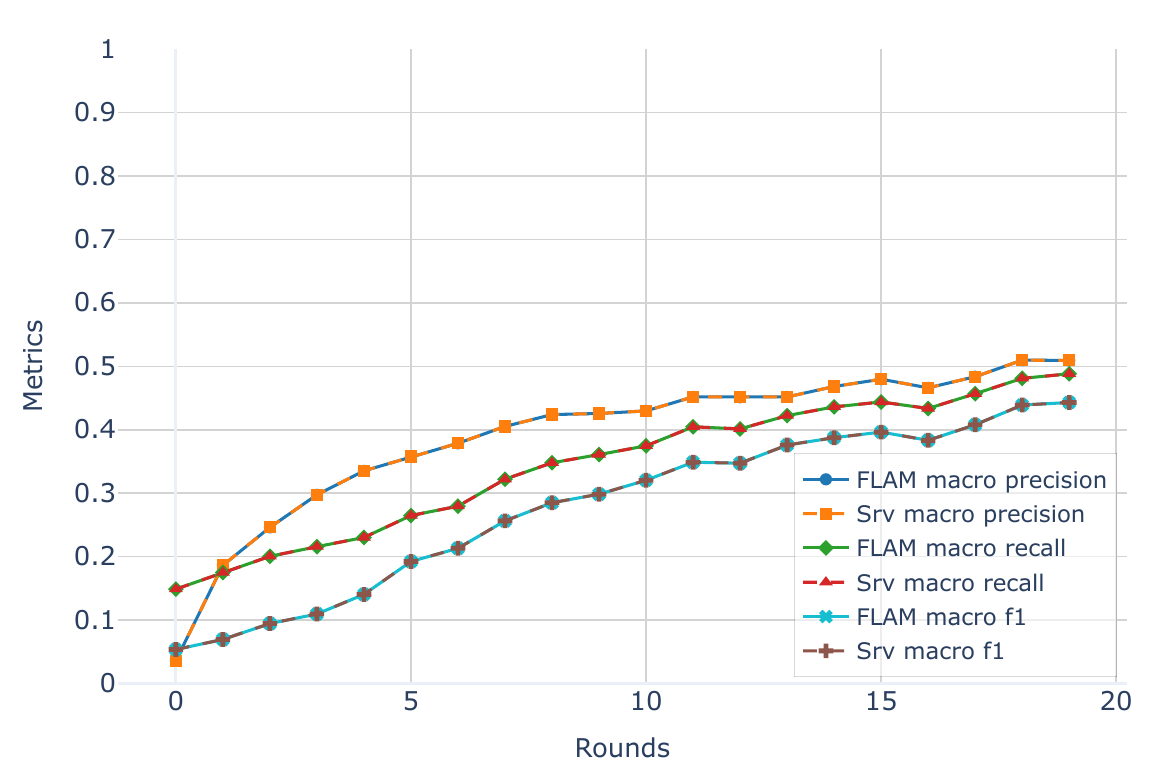}
\caption{Macro Metrics}
\label{fig:solution_exp_cifar_10_ms_mg1}
\end{subfigure}
\begin{subfigure}{195px}
\centering
\includegraphics[width=195px]{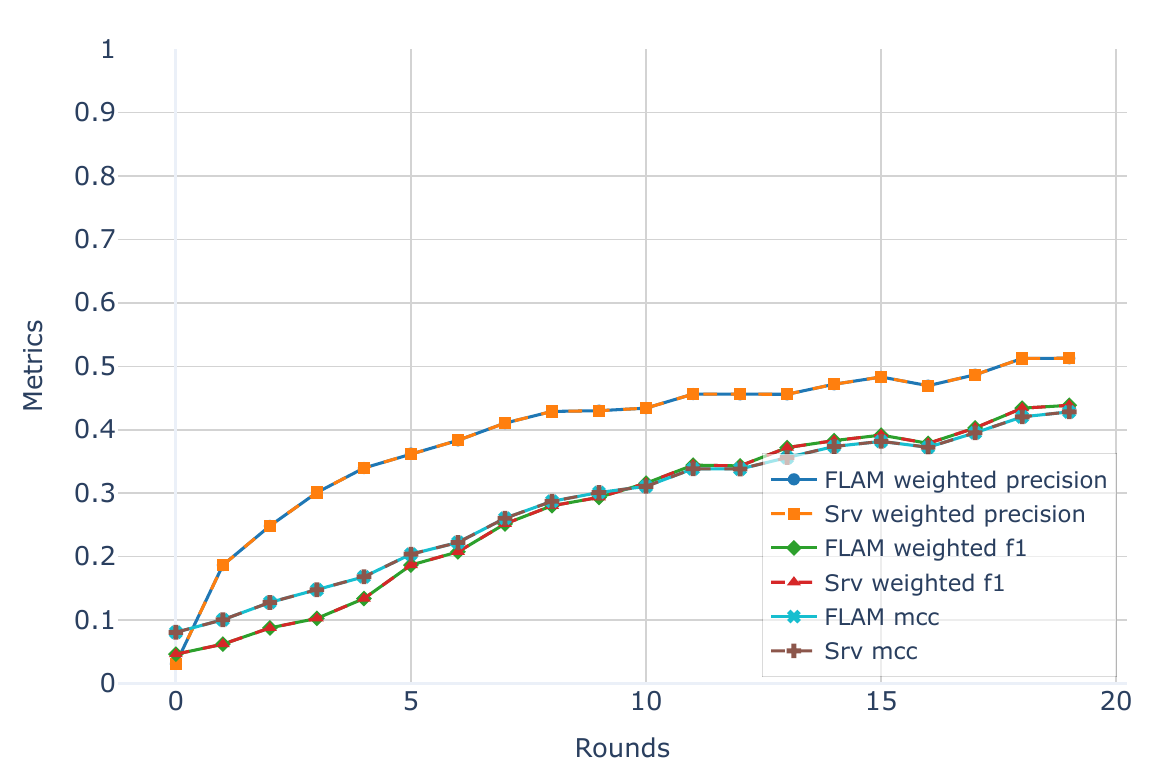}
\caption{Weighted Metrics and MCC}
\label{fig:solution_exp_cifar_10_ms_mg2}
\end{subfigure}
\caption{CIFAR-10 MS: Comparison of FLAM against coordinator-side calculated metrics.}
\label{fig:solution_exp_cifar10_ms_combined}
\end{figure}

\begin{figure}[!t]
\centering
\begin{subfigure}{195px}
\centering
\includegraphics[width=195px]{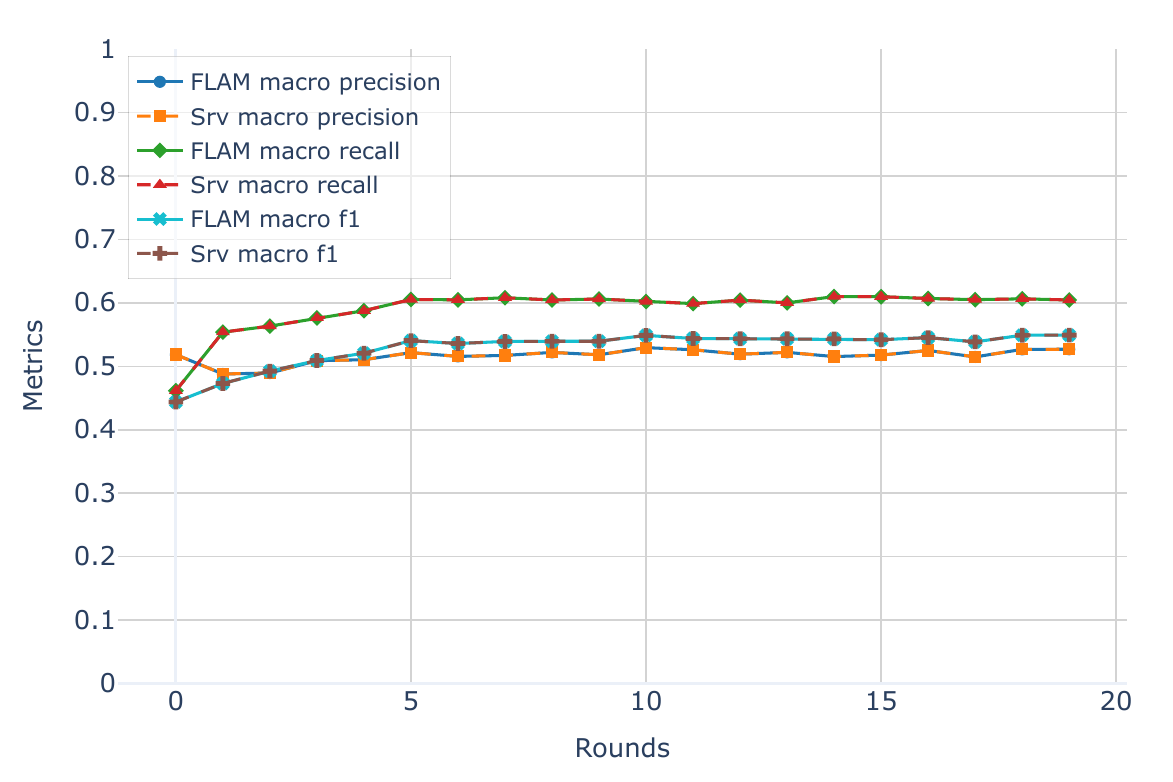}
\caption{Macro Metrics}
\label{fig:solution_exp_covertype_ms_mg1}
\end{subfigure}
\begin{subfigure}{195px}
\centering
\includegraphics[width=195px]{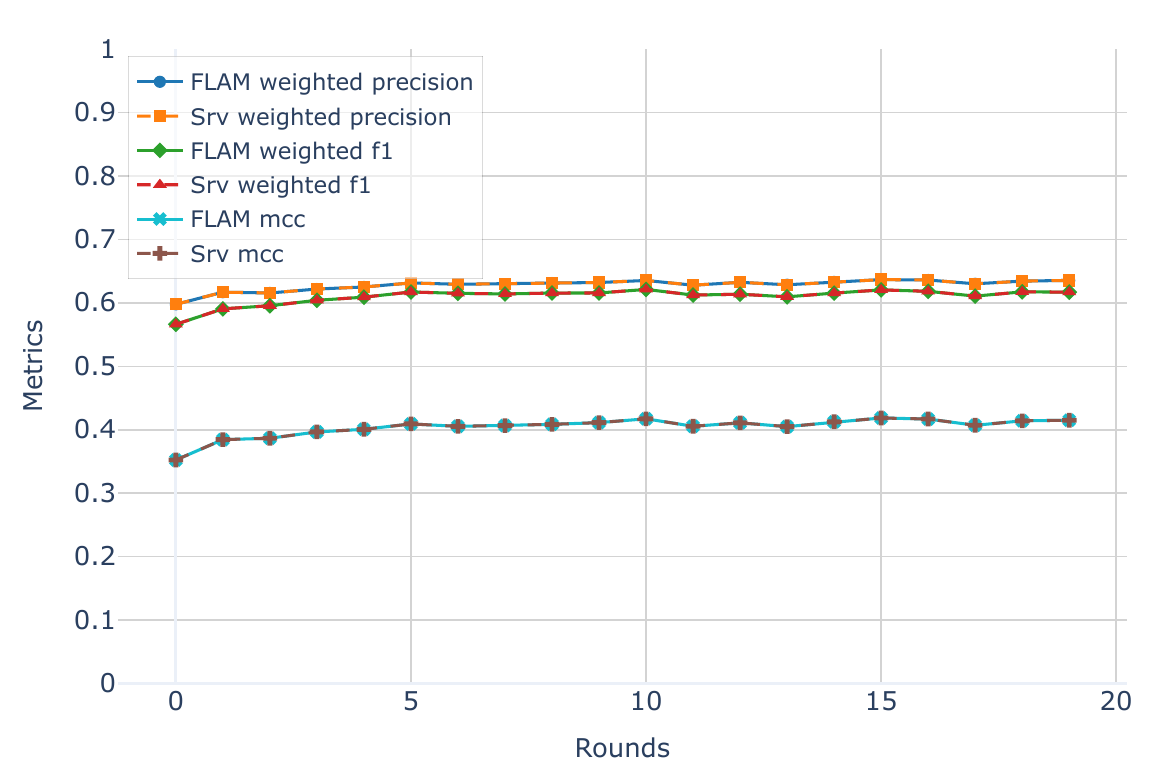}
\caption{Weighted Metrics and MCC}
\label{fig:solution_exp_covertype_ms_mg2}
\end{subfigure}
\caption{Covertype LS with $\alpha=0.6$: Comparison of FLAM against coordinator-side calculated metrics.}
\label{fig:solution_exp_covertype_combined}
\end{figure}

We further show the correctness of the AMs selection by running experiments with the same setup as described in Section~\ref{sec:inacurate_aggregation_of_metrics} and including the evaluation results of FLAM.
Here, we show the results for the experiments that had the highest deviation for the different dataset types.
We start with the regression task in Fig.~\ref{fig:solution_exp_pvod_ms_mg1} that shows that the FLAM achieves the same R2-score as coordinator-based evaluation.
In case of the CIFAR-10 dataset, Fig.~\ref{fig:solution_exp_cifar_10_ms_mg1} indicates the same overlapping results for the macro metrics.
The same applies to the weighted metrics and the MCC in Fig.~\ref{fig:solution_exp_cifar_10_ms_mg2}.
For the Covertype dataset with labels skew and unbalanced classes, the experiments in Fig.~\ref{fig:solution_exp_covertype_ms_mg1} and Fig.~\ref{fig:solution_exp_covertype_ms_mg2} depict, for all metrics, that they achieved the same results as the centralized evaluation.
Following these results, we showed that our method can achieve the same results as centralized evaluation based on a coordinator-side test dataset for commonly used metrics as long as a suitable set of AMs is identified.
Due to the large number of experiments, showing all experiments would be excessive.
We make the experiment results for all metrics publicly available as a GitHub repository.

\subsection{Discussion}
So far, we have shown that our method can achieve the same quality as centralized evaluation for a set of metrics without a centralized test dataset using the participant datasets. Besides the correctness in FL, the applicability and privacy are important aspects.

In FL there are various metrics and learning tasks that we did not include in the evaluation, leading to the question of whether the method can be used with other metrics. 
In general, all metrics are composed of single terms that form the calculation, and the lowest level of AMs in the case of supervised learning would be true and predicted values. These can be used as AMs because these values are counts.
Thus, it would also be possible for any metric in supervised learning that builds upon these values.

Besides the use of different metrics, the method needs to scale with more participants. 
In general, all participants have the same number of measures that they need to calculate, which depends on the metrics and how the equation is decomposed into AMs.
With more participants, only the number of results sent to the coordinator increases, which is more problematic for the aggregation of the model weights, which can be a large matrix.
Here, hierarchical aggregation can be a solution.
Since our method sums up the measures, a hierarchical approach with intermediate aggregators could be a possible solution as long as the metric is calculated on the highest level of the hierarchy of aggregators.
Following the same approach, the computational overhead of summing up the AM of each participant would be distributed alongside the hierarchy of aggregators.

Another aspect is the heterogeneity of participants' data pipelines, which can lead to different data structures and properties. Hence, influencing the training and evaluation.
To achieve a consistent structure, data harmonization is essential for real-world deployment~\cite{kuoResearchCollaborativeLearning2025}. Accordingly, future work should focus on privacy-preserving data harmonization.

Considering privacy, one of the goals of FL is to improve the privacy of the participants and prevent the leakage of sensitive information.
In this regard, sending the true labels and true predictions can lead to privacy issues in real-world FL. 
Although they do not contain the features, in contrast to sending the data to the coordinator, statistical information about the participants' data is revealed.
Therefore, the goal should be to find AMs that consist of multiple aspects in order to reduce information about the data; however, this is strongly dependent on the metric.
While we do not propose privacy solutions, there are methods such as secure multi-party computation (SMPC) to calculate the global mean without revealing individual information of the participants, hence, improving the privacy~\cite{bonawitzPracticalSecureAggregation2017b, soLightSecAggLightweightVersatile2022}.

%% file: sections/conclusion.tex
Evaluating model performance metrics is a common step in ML.
However, in FL, computing the metrics following strategies such as weighted averaging based on the samples per participant can yield different results than centralized evaluation.
These differences can result in incorrect model comparisons and misguided deployment decisions, posing a high risk for the collaboration.

To address this issue, we analyzed the underlying causes of these deviations and showed that the metric calculation is leading to differences.
We proposed FLAM, a general evaluation method that decomposes the calculation into AMs.
These AMs are calculated by the participants and sent to the coordinator to compute the metric.
We showed that this approach achieves the same results as centralized evaluation without the need for a global test dataset.
In addition, we discussed key considerations for reliable model evaluation in FL settings.